\pdfoutput=1

\documentclass[11pt]{article}

\usepackage[]{acl}
\usepackage{times}
\usepackage{latexsym}
\usepackage{booktabs}
\usepackage{multirow}
\usepackage{multicol}
\usepackage{diagbox}
\usepackage{enumitem}
\usepackage{graphicx}
\usepackage{amsmath}
\usepackage{mathtools}
\usepackage{bbm}
\usepackage{bbold}
\usepackage{color, colortbl}
\usepackage{csquotes}
\usepackage[raggedrightboxes]{ragged2e}
\usepackage{url}

\usepackage[T1]{fontenc}

\usepackage[utf8]{inputenc}

\usepackage{microtype}

\definecolor{Gray}{gray}{0.92}
\newcolumntype{g}{>{\columncolor{Gray}}c}
\newcolumntype{P}[1]{>{\centering\arraybackslash}p{#1}}

\title{H\textsc{allu}C\textsc{ana}: Fixing LLM Hallucination with A Canary Lookahead}

\author{{\bf Tianyi Li} \quad {\bf Erenay Dayanik} \quad {\bf Shubhi Tyagi} \quad {\bf Andrea Pierleoni} \\
        Amazon Alexa AI, Cambridge, UK \\
        \texttt{\{tylteddy, erenay, tshubhi, apierleo\}@amazon.co.uk}
}

\begin{document}
\maketitle
\begin{abstract}
In this paper, we present H\textsc{allu}C\textsc{ana}, a canary lookahead to detect and correct factuality hallucinations of Large Language Models (LLMs) in long-form generation.
H\textsc{allu}C\textsc{ana} detects and intervenes as soon as traces of hallucination emerge, during and even before generation.
To support timely detection, we exploit the internal factuality representation in the LLM hidden space, where we investigate various proxies to the LLMs' factuality self-assessment, and discuss its relation to the models' context familiarity from their pre-training.
On biography generation, our method improves generation quality by up to 2.5x, while consuming over 6 times less compute.\footnote{Our code will be released shortly.}
\end{abstract}

\section{Introduction}

The rise of Large Language Models (LLMs) has seen their increasing presence in people's daily lives. Factuality hallucination, a phenomenon where LLMs assertively generate factually unfaithful content \cite{ji_survey_2023,tonmoy_comprehensive_2024}, severely damages users' trust, and blocks their application in critical use cases.

Specifically, for long-form knowledge-intensive LLM responses, mitigating factuality hallucinations w.r.t. world knowledge has attracted increasing attention from the community.
Existing approaches to factuality hallucination detection, such as SelfCheckGPT \cite{manakul_selfcheckgpt_2023} or \citet{varshney_stitch_2023,mundler2024selfcontradictory}, require assistive generations (samples, drafts, etc.), leading to high computational footprint and latency.

In this paper, we propose H\textsc{allu}C\textsc{ana}, \textbf{a novel knowledge-free canary lookahead}, which identifies potential hallucinations selectively and as soon as traces emerge.
H\textsc{allu}C\textsc{ana} is based on \textit{lookahead}, a recent decoding strategy which manipulates generations ad-hoc at the token level, such that repeated re-drafting is avoided.
First introduced by \citet{lu_neurologic_2022} to 
enforce lexical constraints, lookahead has been used to improve faithfulness in abstractive summarization \cite{wan_faithfulness-aware_2023}.

Lookahead evaluates prospective continuations with a scorer, and uses these auxiliary scores to amend the LLM logit scores, so that generation is steered toward desired directions.
For abstractive summarization, prior work \cite{wan_faithfulness-aware_2023} has calculated lookahead scores as entailment between prospective continuations and source documents, where \textit{entails} means \textit{not hallucinated}.
However, for long-form LLM generation w.r.t. world knowledge, no reference documents are directly available.

To overcome this issue, we extract a factuality representation from the LLMs' internal hidden space, as our lookahead scorer.
Following prior work in open-domain QA and factoid statement verification \cite{kadavath_language_2022,azaria_internal_2023}, we train faithfulness classifiers over LLM hidden states, as our lookahead scorer to predict the factuality of continuations.

To ensure generalizability, we explicitly avoid in-domain training sets, and instead train the classifiers on out-of-domain datasets \cite{joshi_triviaqa_2017,kwiatkowski_natural_2019}, then use them off-the-shelf as lookahead scorers in long-form LLM generation. 
Note that we also avoid the complexity and overhead of retrieving external knowledge, so that our approach applies to low-resource domains where reliable knowledge sources are unavailable.

Relevant to using faithfulness classifiers as lookahead scorers, we introduce three innovations to the lookahead. First, since our lookahead inputs are merely hidden states, which do not rely on textual completions, we can predict LLM response factuality pre-hoc, before generation begins.

Second, as faithfulness classifiers are parametric proxies, we observe excessive noise when the classifier is applied at every time step. To address the noise and improve efficiency, we identify \textbf{\textit{critical time steps}} during decoding, so that lookahead is only selectively applied when the LLM is at a crossroad to generating diverse continuations, which are also when hallucination tends to emerge.

For instance, in \textit{King Charles was born in the Buckingham Palace}, it is unambiguous to generate \textit{Palace} after \textit{Buckingham}. Thus, by identifying critical time steps (e.g. \textit{born}, \textit{Buckingham}) and ignoring mere continuations (e.g. \textit{in}, \textit{Palace}), we reduce noise while retaining hallucination detection ability.
We use a logit entropy heuristic to identify critical time steps in decoding, which improves both generation quality and efficiency.

Third, at critical time steps, we observe that branches with high LLM logit scores and low lookahead scores are often still selected during generation, which leads to hallucinations.
Without eclipsing logit scores with too much weight on lookahead scores, we additionally introduce a \textbf{\textit{veto mechanism}}, where branches with very low faithfulness scores are removed, and the choice among remaining branches is left to moderately amended scores.


We conduct evaluation on a long-form generation benchmark generating people's biographies \cite{min_factscore_2023}.
We show that H\textsc{allu}C\textsc{ana}, exploiting LLM internal factuality representation, is robust and grounded in the LLMs' parametric memory of their context familiarity.
In addition to training with accuracy labels, we also train faithfulness classifiers using LLMs' context familiarity from pre-training as supervision labels.
We find that both classifiers predict LLM factual faithfulness in long-form generation comparably well, and that both classifiers produce highly correlated predictions.
On the end-to-end evaluation, H\textsc{allu}C\textsc{ana} improves LLM generation quality by up to 2.5x, outperforming SOTA baselines while consuming over 6 times less compute.

In summary, our contributions include:
\begin{itemize} 
    \item we propose H\textsc{allu}C\textsc{ana}, a light-weight canary lookahead, exploiting the internal factuality representation of LLMs for hallucination detection;
    \item we apply the lookahead selectively at critical time steps, and introduce the veto mechanism in addition to scoring;
    \item we show that this internal factuality representation is grounded in the LLMs' parametric memory of context familiarity, and is robust across natural language tasks.
\end{itemize}

\begin{figure*}[ht]
    \centering
    \includegraphics[width=\linewidth]{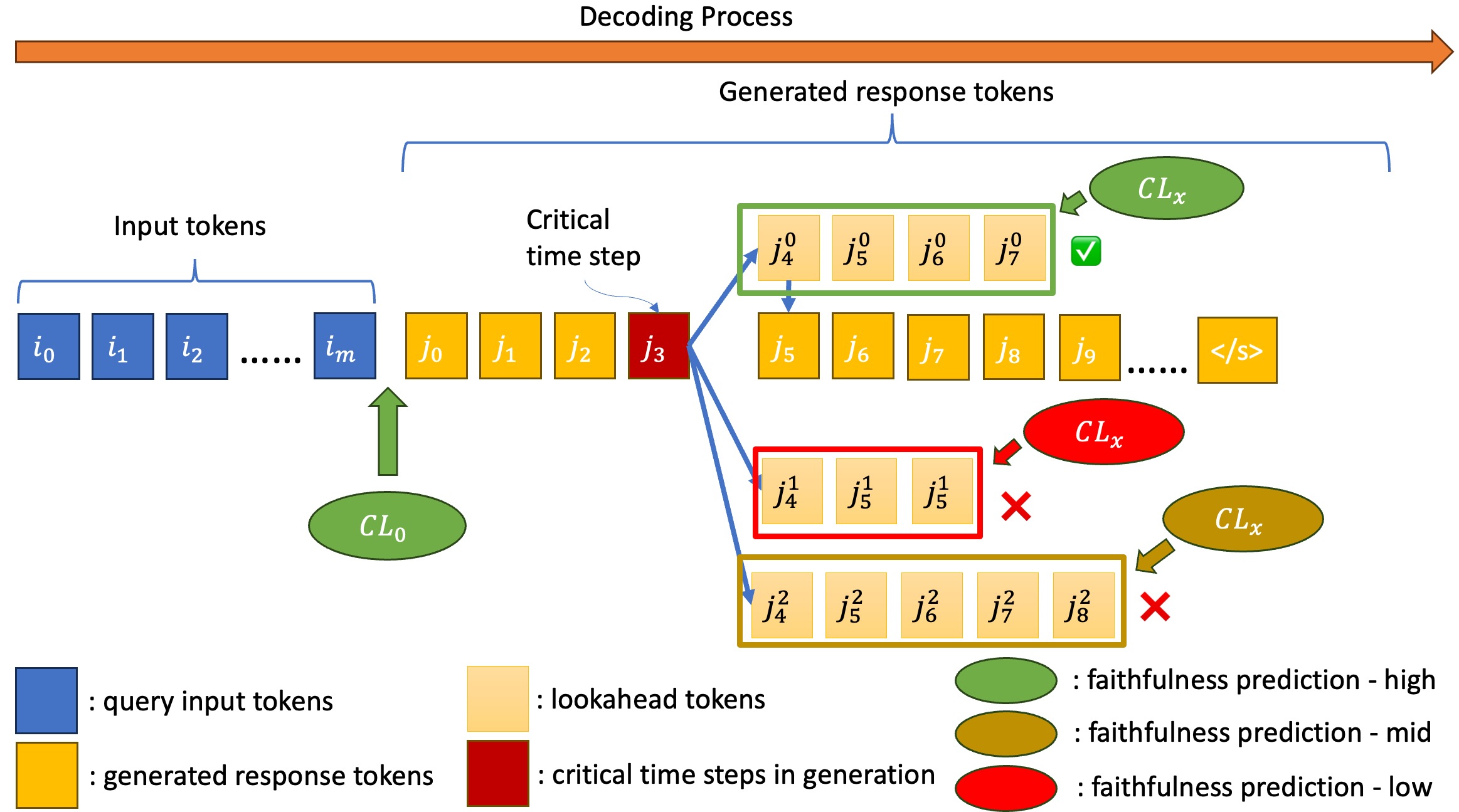}
    \caption{Diagram illustration of H\textsc{allu}C\textsc{ana} in action.}
    \label{fig:Pipeline}
\end{figure*}

\section{Related Work}
\label{sec:related-work}

In abstractive summarization, factuality hallucination has been studied as the inconsistency between summaries and source documents \cite{wan_faithfulness-aware_2023}, and for QA as answer accuracy \cite{kadavath_language_2022,li2023inferencetime}. \citet{maynez_faithfulness_2020} made the distinction between faithfulness w.r.t. source knowledge, and factuality w.r.t. world knowledge. As we are concerned with long-form generation with no specific source \cite{min_factscore_2023,manakul_selfcheckgpt_2023,varshney_stitch_2023}, we simply define hallucination against factuality.

Hallucination detection can be classified into black-box, grey-box, and white-box approaches. Black-box approaches \cite{lin2022teaching,manakul_selfcheckgpt_2023,varshney_stitch_2023,lin_generating_2023,mundler2024selfcontradictory,madaan2023selfrefine,peng_check_2023} only use LLM generated tokens, which apply to closed-source LLMs but require computationally expensive assistive generations. Grey-box approaches \cite{kuhn2023semantic} additionally use logit distributions to help identify hallucinations. 

White-box approaches \cite{kadavath_language_2022,azaria_internal_2023,li2023inferencetime,hernandez_inspecting_2023} develop transformations and classifiers on LLM hidden states, which requires the decoding process to be transparent (i.e. white-box); on the other hand, since LLM hidden states, as the source of both logit distributions and the textual outputs, encodes more fine-grained information, the white-box approach provides a chance to detect hallucinations earlier and more efficiently. 
Our method is also a white-box approach.

When hallucination is detected, several types of mitigation techniques have been used. \citet{kadavath_language_2022} rejects questions with hallucinatory answers; on tasks with fixed answer sets, \citet{wang_self-consistency_2023} samples reasoning paths multiple times and takes the most consistent answer; \citet{mundler2024selfcontradictory} prompts the LLM to delete hallucinatory content and rewrite relevant sentences; \citet{varshney_stitch_2023,madaan2023selfrefine} also re-writes with retrieved knowledge and self-critique, respectively; \citet{li2023inferencetime,hernandez_inspecting_2023} apply hidden state transformations to generate more factual tokens; \citet{chuang2024dola,wan_faithfulness-aware_2023,lee2022factuality} manipulate the LLM decoding process directly to select more factually faithful next-tokens.

Particularly, in abstractive summarization, \citet{wan_faithfulness-aware_2023} uses a decoding strategy called lookahead, first introduced in \citet{lu_neurologic_2022}, to reduce hallucination. We also use a lookahead, but exploit LLMs' internal factuality representation to extract faithfulness lookahead scores instead of external entailment judgements from reference documents. More generally in LLM research, lookahead mechanism has been viewed as a decoding-time alignment strategy to incorporate reward signals, in the form of lookahead scores, into the decoding process \cite{huang2024dealdecodingtimealignmentlarge}.

The LLMs' internal factuality representation has also been discussed in prior work. \citet{kadavath_language_2022,azaria_internal_2023} showed that there exists a representation in the LLM hidden states, which encodes whether a statement or an answer is accurate or not. \citet{yin_large_2023} found that LLMs also possess the ability to detect unanswerable questions. \citet{he-etal-2024-llm} studied patterns in LLM hidden states when generating faithful or hallucinatory answers, using a methodology inspired by physiological lie detectors. On the other hand, \citet{levinstein_still_2023} argues that such representation can be brittle to simple dataset variations; \citet{orgad2024llmsknowshowintrinsic} also shows that this truthfulness representation is different across distinct tasks.
\citet{kossen2024semanticentropyprobesrobust} learnt this truthfulness representation by training on a proxy semantic entropy signal, and observed improvements on generalisabilty.
In this paper, we compare between truthfulness representations learnt from corpus-based familiarity and on accuracy, and establish that for representations learnt from accuracy signals, the generalisable portion of their performance is also grounded in the LLMs' context familiarity.

For more related work on LLM hallucination, we refer the readers to relevant surveys \cite{ji_survey_2023, tonmoy_comprehensive_2024}.

On a separate thread, \citet{kandpal_large_2023,sun_head--tail_2023} have discussed the relation between LLM QA accuracy and the frequency/popularity of involved entities. In this paper, we show that the LLMs are self-aware of their familiarity with contexts, with this information encoded in the LLMs' hidden states; we also show that this awareness is a strong contributor to their factuality awareness.

\section{H\textsc{allu}C\textsc{ana}}
\label{sec:method}

In this section, we introduce our canary lookahead in LLM decoding, as graphically illustrated in Figure \ref{fig:Pipeline}. 
In the below, we first introduce our lookahead architecture in \S\ref{sec:method:lookahead}, then introduce the classifiers in \S\ref{sec:method:preparedness}, where we also discuss the source of factuality classification abilities in LLMs.

\subsection{The Canary Lookahead}
\label{sec:method:lookahead}


Our canary lookahead is designed to reduce hallucination in long-form natural language generation, it consists of two phases: $CL_0$, a pre-hoc scorer, which is applied before generation begins (after $i_m$ and before $j_0$ in Fig. \ref{fig:Pipeline}); $CL_x$, an ad-hoc scorer, which is applied during generation.

When generating a response following an input $i_{0...m}$, we begin by applying the $CL_0$ scorer. $CL_0$ takes the hidden state $hs(i_m)$ generated after encoding the last input token, and predicts the factuality of the not-yet-generated response for input $i_{0...m}$:

\begin{equation*}
    CL_0(i_{0...m})=\textit{classifier}_0(hs(i_m))
\end{equation*}
\vspace{0.0in}

When $CL_{0}(\cdot)<\tau_0$, decoding is not initiated, and an abstained response is returned.


Otherwise, when decoding is initiated, we listen for critical time steps to apply the $CL_x$ scorer.
Critical time steps are decoding steps when the LLM faces uncertainty regarding its subsequent generation. 
We identify these critical time steps by examining the entropy of the LLM logit score distributions. 
For those steps $j_{crit}$, where entropy exceeds a threshold $\tau_{crit}$, $CL_x$ is applied:

\begin{equation*}
    lookahead(j^k_{crit}) = hs(\textit{decode}^{N-1}(j^k_{crit}))
\end{equation*}
\begin{equation*}
    CL_x(j^k_{crit}) = \textit{classifier}_x(lookahead(j^k_{crit})))
\end{equation*}
\vspace{0.05in}

i.e. we create a lookahead branch for each of the top-$K$ likely next token at the critical time step $j^{k}_{crit}$, where we continue greedy decoding ($decode(\cdot)$) for $N$-steps\footnote{The initial token for each branch counts as the first decoded token, so a total of $N-1$ additional decoding steps are applied.}
to allow for a statement to be completed. Then, we take the hidden states at the last lookahead token, $lookahead(j^k_{crit})$, and apply \textit{classifier}$_x(\cdot)$ to predict the faithfulness of this lookahead statement.

We \textit{veto} the branches with $CL_x(\cdot)<\tau_x$, then take weighted geometric mean between the lookahead score (\textit{CL}$_x(\cdot)$) and the LLM logit score (\textit{logit}$(\cdot)$) for the remaining branches to select the next token:
\begin{equation*}
    scr(\cdot) = \begin{cases}
        -\infty , & CL_x(\cdot) < \tau_x \\
        logit(\cdot) \times CL_x(\cdot)^\alpha , & CL_x(\cdot) \geq \tau_x \\
    \end{cases}
\end{equation*}
\begin{equation*}
    k_{selected} = argmax_{k\in K}(scr(j^k_{crit}))
\end{equation*}

where $logit(\cdot)$ is the LLM logit score, $\alpha$ is the weight term. The branch with the highest overall $scr(\cdot)$ is selected. In case all $K$ options are \textit{vetoed}, i.e. $\forall k \in K, scr(t^k)=-\infty$, we simply assign period (.) as the next token to avoid hallucination.

\subsection{Training Faithfulness Classifiers}
\label{sec:method:preparedness}

We now introduce the \textit{classifier}$_0$ and \textit{classifier}$_x$ classifiers corresponding to the two scorers.
While these classifiers are used for long-form generation hallucination detection during inference, we specifically train them on out-of-domain short-form QA datasets to ensure generalisability.
For \textit{classifier}$_0$, we extract hidden states at the last question tokens as training input;
for \textit{classifier}$_x$, we use hidden states at the last answer tokens as training input, reflecting the positions of application during inference.



We explore two types of supervision labels for training classifiers: QA accuracy ($Lbl_{acc}$) and context-familiarity from corpus ($Lbl_{corpus}$).

\paragraph{QA Accuracy} Following prior work \cite{kadavath_language_2022}, we train faithfulness classifiers using string match accuracy of LLM predictions on QA datasets. This is a simple criterion of faithfulness: when the LLM-predicted answer matches the gold answer, in the LLM's internal factuality representation the hidden state should also encode ``factual''.
When the predicted answer matches a gold answer by string match, $Lbl_{acc}=\mathbb{1}$ (factual) is assigned; otherwise, $Lbl_{acc}=\mathbb{0}$ (hallucinatory) is assigned.\footnote{See Appendix \ref{appendix:lbls:acc} for prompts and match metric; we also experimented with NLI-based factuality labels, but did not see improvements.}


\paragraph{Context Familiarity in Corpus} 


Especially for \textit{classifier}$_0$, where no responses or answers are yet specified to be evaluated, and for \textit{classifier}$_x$ as well, we show that the classifiers' ability to detect the factuality of \textbf{their own generations},
is grounded in LLMs' representation for their context familiarity.

Concurred with prior work \cite{kuhn2023semantic}, we notice that hallucination is associated with
LLMs' uncertainty about a statement, which can be traced back to their unfamiliarity with the statement, or more broadly, the statement's context.\footnote{Uncertainty is connected with entropy, however, in long-form generation, high entropy can come from e.g. the uncertainty of either the statement itself or which topic to discuss, so it cannot be used for hallucination detection.}

\citet{kandpal_large_2023} has found that the LLMs' QA accuracy is positively correlated with the frequency of the contexts in LLM pre-train corpora, which embodies context familiarity. 

Inspired by the above, we hypothesize that: 1) from their pre-training, LLMs develop a context familiarity representation in their hidden space; 2) this representation is exploited as a major contributor to the factuality representation, and thus the classifier performance, especially in pre-hoc predictions ($\textit{classifier}_0$).


Based on the hypothesis, directly training on context familiarity should yield comparable or even superior performance.
To verify this, we train classifiers using corpus frequency labels ($Lbl_{corpus}$), which approximates context familiarity.
Importantly, with $Lbl_{corpus}$, we eliminate all traces of factual accuracy from the training process.


Following \citet{kandpal_large_2023}, we use entity combinations to disambiguate contexts; we count their frequencies in LLM pre-train corpora as an upper-bound estimate of LLMs' familiarity with the respective contexts.
For instance, whenever \textit{<Barack Obama, US president, 2008>} are mentioned together, it likely locates the context of \textit{Barack Obama running for office in 2008 and being elected US president.}

For each passage in the corpora, we extract named entities using ReFinED entity linker \cite{ayoola_refined_2022},
and count context mentions by \textbf{consecutive} N-entity spans.
Then, for each QA pair in the training sets, we similarly build N-entity spans and look up their frequencies in the pre-train corpora.
We take geometric mean over these frequencies, as the context-familiarity label.

Notably, we use these frequencies not directly at inference time, but as labels for training. We afford to not assign labels for some entries in the training sets so that noise is reduced for the remaining labels: we restrict N-entity counts to only adjacent mentions, and use 3-entity spans instead of entity pairs (see Appendix \ref{appendix:lbls:corpus} for a comparison).  

\section{Experiment Setup}
\label{sec:experiments:setup}

\paragraph{Models}
In this paper, we focus on two well-known and fully open-source 
LLMs: Falcon-7b-instruct \cite{almazrouei_falcon_2023} and Flan-UL2 \cite{chung_scaling_2022,Tay_2023}.
Falcon-7b-instruct is a strong decoder-only LLM, which makes it especially good at text generation; on the other hand, Flan-UL2 is a strong encoder-decoder LLM widely used for classification tasks.

For the faithfulness classifiers, we use simple 3-layer MLP models, using the last-layer hidden states as inputs. For all our experiments, a frozen falcon-7b-instruct model is used to produce these input hidden states, the LLM parameters are not involved in the training of classifiers.
We refer readers to Appendix \ref{appendix:clsf} for detailed hyper-parameter configs.

\paragraph{Datasets}

Following prior research \cite{min_factscore_2023,manakul_selfcheckgpt_2023, varshney_stitch_2023}, we assess the effectiveness of our canary lookahead in the context of long-form constrained generation, and conduct an evaluation focused on biography generation. Specifically, we use the factscore dataset \cite{min_factscore_2023}: for our development set, we use their ``labelled'' subset, comprising 183 entities, while their ``unlabeled'' subset, containing 500 entities, serves as our test set for reporting the final results.


In order to train the faithfulness classifiers, 
we use two popular generic open-domain QA datasets, Trivia QA \cite{joshi_triviaqa_2017} and Natural Questions \cite{kwiatkowski_natural_2019}, expanding the token-level answers to propositional statements for stylistic consistency.\footnote{For instance, ``Q: Where is the capital of France? A: Paris.'' is expanded to: ``Q: Where is the capital of France? A: The capital of France is Paris.'' See Appendix \ref{appendix:prompts:qa_rephrase}.}
These QA datasets are out-of-domain for the factscore dataset, involving diverse types of question-answer pairs, with less than 0.3\% entity overlap, which we argue is consistent with the natural distribution of text.


When computing corpus frequencies for the context familiarity labels ($Lbl_{corpus}$), we use the respective pre-training corpora for each LLM: for Falcon, we extract frequencies from the falcon-refinedweb corpus \cite{penedo_refinedweb_2023}; for Flan-UL2, we use the C4 corpus \cite{raffel_exploring_2020}.

\paragraph{Hyper-Parameters}
We have introduced 4 hyper-parameters in our canary lookahead method: the lookahead length $N$, the number of explored branches $K$, and the veto thresholds for \textit{CL}$_0$ ($\tau_0$) and for \textit{CL}$_x$ ($\tau_x$).
The lookahead length $N$ is anecdotally set to 8, as a trade-off between allowing atomic statements to finish and avoiding excessive computational cost.\footnote{When sentence-end is met before looking N tokens ahead, the lookahead decoding also stops since the statement has already finished.}
The number of branches is set to $K=3$ as a minimum default value, where we already observe substantial gains.
On the other hand, the two veto thresholds are empirically selected to maximize development-set F-scores along the F-R curves (to be elaborated in the below). Specifically, for \textit{CL}$_x$ scorers, since \textit{veto} is repeatedly applied through a decoding run, we assign more benign $\tau_x$ maximizing F-1.2; for \textit{CL}$_0$ and baselines, F-1 is maximized.


\paragraph{Baselines}
We compare our approach against various baselines: 1) vanilla LLMs with greedy and beam decoding (using beam size of 4 for comparable computational footprint); 2) \textit{SelfCheckGPT} (\textit{w/NLI} variant, as recommended) \cite{manakul_selfcheckgpt_2023}, current SOTA of hallucination detection; 3) MAX($-logp$), used as a strong baseline in \citet{manakul_selfcheckgpt_2023}.\footnote{Apart from these, \citet{varshney_stitch_2023} is also relevant to our work. We exclude this baseline for two reasons: 1, it assumes the availability of external knowledge; 2, no implementation is available for its reproduction.}
For \textit{SelfCheckGPT}, sample responses are generated for reference, then response factuality is predicted using its consistency with the generated references.
For MAX($-logp$), we calculate the minimum logit probability of generated tokens over a piece of response, and use it as a baseline measure of the LLMs' self-assessed uncertainty.

\paragraph{Metrics}
To compare H\textsc{allu}C\textsc{ana} against baselines, we measure the factuality of generated responses. We use an automatic measure, FActScore (FS) \cite{min_factscore_2023}, which judges whether each atomic fact in the response is supported by a gold KG.\footnote{We use a SOTA Mistral LLM \cite{jiang_mistral_2023} in FActScore calculation, which yields consistent results as the original implementation, see Appendix \ref{appendix:metrics:factscore} for details.} For each query-response pair, FS produces a floating point value between [0, 1], which denotes the ratio of supported atomic facts in the response.


For \textbf{evaluating classification performance} (\S\ref{sec:experiments:classifiers}),
since the FS labels are non-binary, we are unable to use the P-R curves. Instead, we introduce the FS-Rejection-curve, similar to the Accuracy-Rejection-curve in \citet{lin_generating_2023} (see Appendix \ref{appendix:metrics:aufrc} for graphic explanations).
We calculate \textit{AUC-FR}, Area Under the FS-Rejection Curve, as our classification metric; we normalize the values between random classifiers $\textit{AUC-FR}_{norm}(\textit{rand})=0$ and perfect classifiers $\textit{AUC-FR}_{norm}(\textit{pft})=1$.
Intuitively, $\textit{AUC-FR}_{norm}$ measures the ratio of performance realized by the classifier, relative to its theoretical upper bound:

\begin{equation*}
    \textit{AUC-FR}_{norm}=\frac{\textit{AUC-FR}-\textit{AUC-FR}(\textit{rand})}{\textit{AUC-FR}(\textit{pft})-\textit{AUC-FR}(\textit{rand})}
\end{equation*}

\begin{table*}[t]
    \centering
    \begin{tabular}{c|c|g}
    \toprule
    \textbf{Method} & \textbf{Phase} & \boldmath{$\textit{\textbf{AUC-FR}}_{norm}$} \\
    \midrule
    \textit{MAX($-logp$)} & $end$ & \textit{36.3} \\ 
    \textit{SelfCk \cite{manakul_selfcheckgpt_2023}*} & $end$ & \textit{52.6*} \\
    \textit{classifier}$_0$-$Lbl_{acc}$ \cite{kadavath_language_2022} & $CL_0$ & 44.2 \\
    \textbf{\textit{classifier}}\boldmath{$_0$}\textbf{-}\boldmath{$Lbl_{corpus}$} & \boldmath{$CL_0$} & \textbf{48.1} \\ 
    \midrule
    MAX($-logp$) & $CL_x$ & 10.4 \\
    \textit{SelfCk \cite{manakul_selfcheckgpt_2023}*} & $CL_x$ & \textit{48.8*}\\
    \textbf{\textit{classifier}}\boldmath{$_x$}\textbf{-}\boldmath{$Lbl_{acc}$} \cite{kadavath_language_2022} & \boldmath{$CL_x$} & \textbf{37.5} \\ 
    \textit{classifier}$_x$-$Lbl_{corpus}$ & $CL_x$ & 23.1 \\
    \bottomrule
    \end{tabular}
    \caption{Classification performance for falcon-7b-instruct LLM, measured by \% of $AUC{-}FR_{norm}$ on greedy generations for \textbf{factscore dev set}. \textit{SelfCk}\shortcite{manakul_selfcheckgpt_2023}* requires generating extra reference responses, thus the asterisk. $CL_0$ means prediction is made before decoding, $CL_x$ means during decoding at sentence-ends, $end$ means after generation ends.} 
    \label{tab:preparedness_classifier}
\end{table*}



For \textbf{hallucination reduction performance} (\S\ref{sec:experiments:lookahead}), we use FActScore for the not-abstained responses, as the factuality metric. In addition, we measure informativeness, using the number of atomic facts (\textit{\#Facts}$_d$) in each response $d$, and responsiveness, using the ratio of not-abstained responses (\textit{\% respond}).\footnote{We count a response as \textit{abstained}, when it starts with ``sorry'', or is shorter than 16 words, or is predicted as \textit{hallucinatory} at $CL_0$.}

We combine these aspects of generation quality with a \textbf{\textit{Generation Quality Index}} ($GQI$):
\begin{equation*}
    gqi_{\gamma}(d) = (\textit{FS}_d-\gamma)\times \textit{\#Facts}_d
\end{equation*}
\begin{equation*}
    GQI_{\gamma}=\frac{\sum_{d\in D}gqi_{\gamma}(d)}{|D|\times \sqrt{\textit{response rate}}}
\end{equation*}

$GQI$ reflects our belief in response values: 1) responses with $FS_d < \gamma$ are negative assets, thus the $\textit{FS}_d-\gamma$ term (we empirically set $\gamma=0.2$ for reporting results in \S\ref{sec:experiments:results}); 2) longer responses for less queries are preferred over short responses for more queries, thus the $\sqrt{\cdot}$.
We use $GQI$ as a holistic metric for straightforward comparison, but we also provide individual metrics for finer-grained inspections.
We include a detailed analysis of the $GQI_{\gamma}$ metric in Appendix \ref{appendix:metrics:gqi}.


Apart from generation quality, we also measure the computational footprint, 
using the average number of tokens 
processed by the LLM, for an atomic fact to be generated in the final response, namely, \textbf{\textit{Tokens per Fact}} ($TpF$):
\begin{equation*}
    TpF=\frac{\sum_{d\in D}{(\lambda*|seq^{enc}_d| + |seq^{dec}_d|)}}{\sum_{d\in D}{\textit{\#Facts}_d}}
\end{equation*}
where the $0<\lambda<1$ reflects that token encoding cheaper than token decoding. Specifically, we set $\lambda=0.5$, which is aligned with the pricing ratio with Open AI (\url{https://openai.com/pricing}, visited Jan 2, 2024). For simplicity, we ignore the computational footprint our classifiers, which are light-weight and called only once per explored branch, as well as smaller NLI models in \textit{SelfCheckGPT} baselines.

\begin{table*}[t]
    \centering
    \begin{tabular}{c|g|ccc|g}
    \toprule
    \textbf{Method} & \boldmath{$GQI_{0.2} \uparrow$} & \textbf{Avg. FS (\%) $\uparrow$} & \textbf{Avg. \#Facts $\uparrow$} & \textbf{\% respond $\uparrow$} & \boldmath{$TpF \downarrow$} \\
    \midrule
    greedy & 0.68 & 25.4 & 15.2 & 100\% & 7.8 \\ 
    beam-4 & 1.01 & 29.9 & 11.7 & 94.2\% & 53.4 \\ 
    \textit{MAX($-logp$)} & 1.07 & 29.7 & 14.4 & 69.1\% & 12.0 \\ 
    \textit{SelfCk*} \shortcite{manakul_selfcheckgpt_2023} & 1.36 & 32.6 & 15.3 & 66.9\% & 298.5 \\ 
    \midrule
    \multicolumn{6}{l}{\textbf{H\textsc{allu}C\textsc{ana}}} \\
    \cmidrule{2-6}
    $CL_0$ & 1.17 & 30.3 & 15.1 & 71.7\% & 7.9 \\ 
    $CL_x$ & 1.34 & 30.9 & 14.0 & 75.4\% & 51.6 \\ 
    \boldmath{$CL_0+CL_x$} & \textbf{1.69} & \textbf{36.0} & \textbf{14.4} & \textbf{56.3\%} & \textbf{47.1}  \\ 
    $(CL_x)_{corpus}$ & 1.07 & 30.8 & 10.5 & 94.2\% & 48.2 \\ 
    $(CL_0+CL_x)_{corpus}$ & 1.46 & 33.6 & 12.5 & 57.3\% & 46.5 \\ 
    
    \bottomrule
    \end{tabular}
    \caption{Hallucination reduction performance for falcon-7b-instruct on \textbf{factscore test set}. For strict compute budgets, $CL_0$ is good for preventing hallucination; generally, full H\textsc{allu}C\textsc{ana} ($CL_0 + CL_x$) (\textbf{bolded}) is recommended.}
    \label{tab:decoding-falcon7b}
\end{table*}

\section{Results and Discussions}
\label{sec:experiments:results}

We report results on the stand-alone classification task in \S\ref{sec:experiments:classifiers}, and end-to-end on long-form generation in \S\ref{sec:experiments:lookahead}. In the main tables are results with Falcon-7b-instruct, we leave results with Flan-UL2 to App. \ref{appendix:flanul2}, where conclusions are consistent.

\subsection{Faithfulness Classifiers}
\label{sec:experiments:classifiers}

In Table \ref{tab:preparedness_classifier}, we report our faithfulness classifier's stand-alone classification performance, and compare it with baselines described in \S\ref{sec:experiments:setup}.

We measure how well classifier predictions align with FActScore labels on the LLM greedy generations on factscore dev set.
We evaluate each classifier against what they predict: we evaluate $\textit{classifier}_0$ classifiers and $end$ baselines against the factuality of full responses, and $\textit{classifier}_x$ against that of the current sentences ending with the classifier input tokens.

In general, faithfulness classifiers (\textit{clsf-}), trained on QA datasets, generalize well to hallucination detection in long-form generation: they exhibit non-trivial performance, achieving up to half the performance of a perfect classifier, consistently outperforming the logit baseline (\textit{MAX($-logp$)}) (48.1 vs. 36.3; 37.5 vs. 10.4). Especially in the $CL_0$ phase, they achieve results comparable to the far more expensive \textit{SelfCk} baseline, without requiring tokens to be decoded ( $CL_0$ vs. $end$ ).


Different training signals are favoured in $CL_0$ and $CL_x$ phases. $CL_0$, the pre-hoc phase, depends more strongly on context familiarity; as expected, training directly on context familiarity ($Lbl_{corpus}$) yields superior results to QA accuracy ($Lbl_{acc}$) (48.1 vs 44.2).
This supports our hypothesis in \S\ref{sec:method:preparedness}, that when we train on QA accuracy labels, we end up learning a similar representation as when training on context familiarity labels. 

Moreover, predictions from classifiers trained on the two signals are also strongly correlated, with Spearman's $\rho(Lbl_{acc}, Lbl_{corpus})=0.63$, and p-value 1.4e-21. For comparison, between predictions from the \textit{SelfCk} baseline and these two classifiers, this metric are $\rho(\textit{SelfCk}, Lbl_{acc})=0.37$ and $\rho(\textit{SelfCk}, Lbl_{corpus})=0.31$, respectively, both much lower.


On the other hand, for $CL_x$, the ad-hoc phase, \textit{classifier}$_x$-$Lbl_{acc}$ prevails. This shows that at the level of individual statements, other aspects of the faithfulness representation than context familiarity, such as distractors, etc., are non-negligible and also encoded in LLM hidden space, but not captured in context familiarity. Nevertheless, without any exposure to factual accuracy, \textit{classifier}$_x$-$Lbl_{corpus}$ still achieves non-trivial performance.




As the final verdict, we take \textit{classifier}$_0$-$Lbl_{corpus}$ for $CL_0$ and \textit{classifier}$_x$-$Lbl_{acc}$ for $CL_x$ as the default setup for our H\textsc{allu}C\textsc{ana} lookahead to be evaluated below, and additionally report results solely using classifiers trained on $Lbl_{corpus}$, as a robustness lower-bound.



 

\subsection{The Canary Lookahead}
\label{sec:experiments:lookahead}

We now evaluate H\textsc{allu}C\textsc{ana} end-to-end, on reducing hallucination in long-form LLM generation.

When applying the lookahead, we convert classifier predictions to binary verdicts using representative thresholds on the dev set, maximizing F-scores along the F-R curves as elaborated in \S\ref{sec:experiments:setup}. 


Table 2 presents the results.
First, when we use pre-hoc scorers to prevent hallucination before decoding ($CL_0$), we approach the performance of the SOTA \textit{SelfCk} baseline, while requiring less than 3\% of its compute (7.9 vs. 298.5).

Second, when we use ad-hoc scorers to manipulate decoding towards faithful continuations ($CL_x$), we outperform beam decoding: we get 1.0 point higher FActScore, and generate 20\% more atomic facts per response trading off 20\% responses.

\begin{table*}[t]
    \centering
    \begin{tabular}{c|g|ccc|g}
    \toprule
    \textbf{Ablation} & \boldmath{$GQI_{0.2} \uparrow$} & \textbf{Avg. FS (\%) $\uparrow$} & \textbf{Avg. \#Facts $\uparrow$} & \textbf{\% respond $\uparrow$} & \boldmath{$TpF \downarrow$} \\
    \midrule
    \boldmath{$CL_0+CL_x$} & \textbf{1.52} & \textbf{33.7} & \textbf{14.8} & \textbf{48.6\%} & \textbf{48.7} \\
    \cmidrule{2-6}
    \multicolumn{1}{r|}{$-$ \textit{critical}} & 0.70 (\textcolor{purple}{-0.82}) & 36.4 (\textcolor{teal}{+2.7}) & 10.5 (\textcolor{purple}{-4.3}) & 16.4\% (\textcolor{purple}{-32.2\%}) & 133.4 (\textcolor{purple}{+84.7}) \\ 
    \multicolumn{1}{r|}{$-$ \textit{veto}} & 0.63 (\textcolor{purple}{-0.89}) & 25.3 (\textcolor{purple}{-8.4}) & 19.7 (\textcolor{teal}{+4.9}) & 66.7\% (\textcolor{teal}{+18.1\%}) & 50.9 (\textcolor{purple}{+2.2}) \\ 
    \multicolumn{1}{r|}{$-$ \textit{score}} & 0.90 (\textcolor{purple}{-0.62}) & 32.9 (\textcolor{purple}{-0.8}) & 14.1 (\textcolor{purple}{-0.7}) & 49.7\% (\textcolor{teal}{+1.1\%}) & 48.8 (\textcolor{purple}{+0.1}) \\ 
    \multicolumn{1}{r|}{$-$ \textit{continuation}} & 1.44 (\textcolor{purple}{-0.08}) & 40.9 (\textcolor{teal}{+7.2}) & 16.9 (\textcolor{teal}{+2.1}) & 33.3\% (\textcolor{purple}{-15.3\%}) & 12.8 (\textcolor{teal}{-35.9}) \\ 
    \midrule
    beam-4 & 0.67 \textcolor{purple}{-0.85} & 26.7 (\textcolor{purple}{-7.0}) & 10.5 (\textcolor{purple}{-4.3}) & 97.8\% (\textcolor{teal}{+49.2\%}) & 55.5 \textcolor{purple}{+6.8} \\ 
    \bottomrule
    \end{tabular}
    \caption{Ablation study over \textbf{H\textsc{allu}C\textsc{ana}}, for Falcon-7b-instruct generated biographies on \textbf{factscore dev set}.}
    \label{tab:decoding-falcon7b-ablation}
\end{table*}

Crucially, our full lookahead ($CL_0+CL_x$) yields the best overall generation quality, improving over vanilla greedy decoding by up to 2.5x; we also improve over prior SOTA by 24\%, while consuming 6 times less compute (47.1 vs. 298.5).

Interestingly, using classifiers trained with $Lbl_{corpus}$ context-familiarity labels without exposure to accuracy, $(CL_x)_{corpus}$ and $(CL_0+CL_x)_{corpus}$ also effectively reduces hallucinations and improves generation quality, both outperforming their respective baselines. This shows the key role of context familiarity in the classification performance and the robustness of H\textsc{allu}C\textsc{ana}.


To summarize, we may conclude:
\begin{enumerate} 
    \item the $CL_0$ scorer is an effective non-invasive hallucination detection technique, which is always recommended in application;
    \item where compute budget allows, the full lookahead is recommended, as it substantially improves factual faithfulness of LLM generations, while being informative and responsive;
    \item the LLM internal representation of context familiarity is a major contributor to the factuality classification performance, such that classifiers trained on context familiarity are able to effectively detect hallucination in long-form generation;
\end{enumerate}

\subsection{Ablation Studies}
\label{sec:experiments:ablation}

We conduct ablation studies on the factscore dev set to evaluate our design choices. Table \ref{tab:decoding-falcon7b-ablation} presents the results.

\paragraph{Is it important to identify Critical Time Steps?} 
\textbf{\textit{Yes}}. 
As introduced in \S\ref{sec:method:lookahead}, we use an entropy-based heuristic to identify \textbf{critical time steps}, and apply $CL_x$ scorer only at these time steps. 

In the $-$ \textit{critical} row, we report results without identifying critical time steps. The average FS is higher, but responses are substantially shorter (-4.3 atomic facts per response), and the response rate suffers severely (48.6\% $\rightarrow$ 16.4\%). This shows, that selectively applying $CL_x$ at critical time steps is essential for reducing noise and false vetos, and does not spoil response factuality.\footnote{Note that the first few facts in a biography tend to be easier to infer and have higher FActScores (e.g. inferring nationality from names, etc.), therefore with the large difference in average length and responsiveness, the moderate difference in \textbf{Avg. FS} is not sufficiently indicative.}
In addition, critical time steps also help reduce the computational footprint ($TpF$ down 133.4 $\rightarrow$ 48.7).

\paragraph{Is it important to veto continuations with low \boldmath{$CL_x$} scores?}
\textbf{\textit{Yes}}.
When the ad-hoc lookahead score $CL_x(j^k_{crit}) < \tau_x$, we veto the branch $j^k_{crit}$ because it likely leads to hallucination.

In the $-$\textit{veto} row, we report results without veto. With the $CL_x$ scores, the LLM generates more information before hitting <eos> (+4.9 atomic facts per response), but the average FActScore of these responses drops sharply (33.7 $\rightarrow$ 25.3). This is because, without the veto mechanism, $CL_x$ cannot help when all tested options are hallucinatory. 

\paragraph{Is it important to amend logit scores in addition to vetoing?}
\textbf{\textit{Yes}}. When the $CL_x$ score is above threshold $\tau_x$ (i.e. not vetoed), we take the weighted geometric mean between $CL_x$ and logit scores as the final score to select the next tokens.

In the $-$\textit{score} row, we remove this score amendment, and observe a significant drop in generation quality. Note that \textbf{Avg. FS} takes a macro-average over the responses, whereas $GQI$ incorporates a micro-average. $GQI$ index suffers more than \textbf{Avg. FS} and \textbf{Avg. \#Facts}, this means, $CL_x-$\textit{score} leads to a skewed distribution of FActScores where long responses have much lower FS than short ones.

Summarizing the two ablations above, for $CL_x$, the veto mechanism is important for purging hallucinatory continuations, and the scoring mechanism is also important, for keeping the LLM on track of having faithful contents to present.  

\paragraph{Can we further improve efficiency by avoiding decoding the continuations for $CL_x$?} \textbf{\textit{Yes and no.}}
Also with the $CL_x$ scorer, from each likely branch $j^k_{crit}$, we decode for up to $N=8$ steps to complete the current statement before predicting factuality. If we avoid these extra decoding steps, and apply the classifier after the first lookahead token (effectively setting $N=1$), $CL_x$ would become more lightweight, with negligible computational overhead.

In the $-$\textit{continuation} row, we report lookahead performance without the extra decoding steps. We observe relatively promising results with generation quality only slightly lower than our main setup. Especially, the average FActScore and the average number of atomic facts are both higher; however, the con side is that its response rate is excessively low, where only a third of queries receive valid responses. Even without the $CL_0$ filtering, the ratio of valid responses is still only 41.5\%.

Therefore, we do not use this config as our main setup, but note that users with low response rate expectations and/or tight computation constraints (the $TpF$ metric for $-$\textit{continuation} is comparable to that of regular greedy decoding) are still encouraged to apply $tok\_x$ without extra decoding.



\section{Conclusion}
In this paper, we introduce H\textsc{allu}C\textsc{ana}, a canary lookahead to address factuality hallucination in LLM long-form generation. 
H\textsc{allu}C\textsc{ana} consists of two phases: a pre-hoc scorer applied before generation begins, and an ad-hoc scorer applied during generation at critical decoding steps.

H\textsc{allu}C\textsc{ana} detects hallucinations using lightweight classifiers trained using out-of-domain data to extract the internal factuality representation in the LLM hidden space. Through our experiments, we show that H\textsc{allu}C\textsc{ana} substantially outperforms prior SOTA methods on generation quality, while at the same time consuming 6x less compute. Furthermore, our exploration reveals the intimate correlation between the LLM's internal representations for factuality and the familiarity of context. The performance of the faithfulness classifiers is deeply rooted in the parametric memory of context familiarity, embedded within the LLMs and acquired during the pre-training phase.

For future work, we plan to improve context disambiguation for context familiarity proxy, explore more sophisticated critical time step identifiers, and extend H\textsc{allu}C\textsc{ana} to reasoning-intensive tasks.


\section*{Limitations}

In this paper, our experiments are based on the benign users' assumption, where we do not consider prompt attacks.

Also, we develop H\textsc{allu}C\textsc{ana} only based on greedy decoding. We leave experiments with best-first-search to future work.

Our evaluation metric focuses on factuality, informativeness and responsiveness, where we do not quantitatively evaluate the fluency of the responses due to the lack of an automatic metric. We offer a case study in Appendix \ref{appendix:case_study} where we qualitatively compare the generation fluency between our method and baselines.

Our approach, falling under the category of white-box approaches, does not apply to LLMs with closed-source model parameters. For the benefit of the analysis, we use LLMs whose pre-train corpora are also available, although we note that in the absence of the LLMs' own pre-train corpora, other open-source large pre-train corpora such as falcon-refiendweb or C4, are also good for inducing the context-familiarity proxy. 

Our evaluation metric for computational footprint does not differentiate the amount of compute and latency. This is because we assume tight computing resources, where the overall objective is not only to reduce latency for individual queries, but also to reduce the overall workload of serving.


\bibliography{anthology,references_tianyi,references_manual}
\bibliographystyle{acl_natbib}

\appendix

\section{Prompts in LLM generations}
\label{appendix:prompts}

\subsection{Q-A Rephrasing}
\label{appendix:prompts:qa_rephrase}

For stylistic consistency, we expand the answers in the QA data entries into sentences. To do so, we use the Mistral-7B-Instruct-v0.1 model \cite{jiang_mistral_2023}, and prompt it to take in a Q-A pair and convert it into a declarative sentence. We use one in-context example in the prompt, and wrap the prompt in a chat template provided with Mistral:

\begin{displayquote}
    \textbf{User:} \textit{Paraphrase the following Q-A pair into a proposition. Q: Where is the capital of France? A: XXXX.}
    
    \textbf{Assistant:}  \textit{The capital of France is XXXX.}

    \textbf{User:} \textit{Paraphrase the following Q-A pair into a proposition: Q: \textsc{[question]} A: XXXX.}

    \textbf{Assistant:}
\end{displayquote}

When training the faithfulness classifiers (specifically, the $\textit{classifier}_x$ classifiers), hidden states at the last expanded answer tokens are used as training input. 
The placeholder ``XXXX'' is installed here, so that it can be easily replaced with different answer tokens:
when training using $Lbl_{acc}$ labels, LLM predicted answers are plugged in to model the factuality of this predicted answer; when using $Lbl_{corpus}$ labels, gold answers from the training dataset are used to purge exposure to accuracy, and model only context familiarity.

\subsection{Long-form Generation}
\label{appendix:prompts:longform}

When generating biographies for the factscore dataset, we use the following conversational prompt:

\begin{displayquote}
    \textbf{User:} \textit{Can you write me a bio to introduce \textsc{[person]}?}
    
    \textbf{Assistant:}  \textit{Sure, here's their bio\textcolor{red}{:} }
\end{displayquote}

Note that for the $CL_0$ scorer in the lookahead, we apply it at the last prompt token, i.e. the ``\textcolor{red}{:}'' after ``bio'' in the prompt response.

\section{Labels for Faithfulness Classifiers}
\label{appendix:lbls}

\subsection{$Lbl_{acc}$ Details}
\label{appendix:lbls:acc}

When training $\textit{classifier}_x$ faithfulness classifiers with $Lbl_{acc}$ labels, we extract hidden states using LLMs' own predicted answers. These answers are predicted using greedy decoding, with a set of four few-shot examples, wrapped in a simple Q-A template. Specifically:

\begin{displayquote}
    \textit{Q: \textsc{[example 1]} A: \textsc{[example 1]}}

    \textit{Q: \textsc{[example 2]} A: \textsc{[example 2]}}

    \textit{Q: \textsc{[example 3]} A: \textsc{[example 3]}}

    \textit{Q: \textsc{[example 4]} A: \textsc{[example 4]}}

    \textit{Q: \textsc{[example 5]} A: \textsc{[example 5]}}

    \textit{Q: \textsc{[Question]} A: }
\end{displayquote}

Accuracy labels are then acquired by string match between the predicted answer and the gold answer. A string match is met when the predicted answer exactly matches the gold answer, or when the predicted answer is a prefix or a postfix of the gold answer, and vice versa.

Notably, we also tried using NLI labels produced by DeBERTa-V3 \cite{he_debertav3_2023} to replace string match as the measure of factual accuracy. However, using NLI labels yielded consistently inferior results, therefore we leave them out of this paper.

\subsection{$Lbl_{corpus}$ Details}
\label{appendix:lbls:corpus}

When training $\textit{classifier}_0$ and $\textit{classifier}_x$ classifiers with context corpus-familiarity labels ($Lbl_{corpus}$), we use the frequency of contexts in pre-training corpora as the proxy.

We extract mentions of named entities and mentions of time, year, ordinals, etc., with the ReFinED entity linker \cite{ayoola_refined_2022}, and acquire entity spans from them.

\citet{kandpal_large_2023} used entity combinations that co-occur in the same passage, as the criterion for context mentions in pre-train corpora. We observe that this criterion is noisy when entity pairs are considered co-occurring whenever they both occur in the same document: they could be far apart and have very weak semantic connections. Therefore, we impose the restriction where only adjacently-appearing entities are considered co-occurring.

Differently from \citet{kandpal_large_2023,sun_head--tail_2023}, who analyse the relation between entities frequencies/popularities and the LLMs' parametric knowledge, our goal here is to extract an internal context-familiarity representation from the LLMs' own hidden space. Therefore, we do not need to assign a corpus-frequency label to every entry, as long as the labelled entries to be included in the training set are sufficiently representative. 

As such, in addition to the adjacency restriction mentioned above, we also define ``context'' in a more restrictive, finer-grained way. We count two mentions in the pre-train corpora as mentions of the same context, only when the same set of 3 entities appear consecutively in both passages. This is also more stringent than the entity pairs, which leads to less coverage, but also reduces noise and yields more accurate contexts. For instance, the entity pair \textit{<France, Argentina>} is very broad and ambiguous, whereas the 3-tuple \textit{<France, Argentina, world cup>} is much more precise, where all the mentions with these entities appearing consecutively together are truly relevant to each other.

Without blindly trusting the string-match accuracy of LLM predictions as the gold metric of success, we do argue that an excessively low correlation between the context familiarity labels and string-match accuracy labels indicate excessive noise. Thus, in Table \ref{tab:lblcorpus}, we validate our choice of $Lbl_{corpus}$, by comparing among different instantiations of $Lbl_{corpus}$ with respect to this correlation. From the results, we observe that \textbf{consecutive 3-gram} is the least noisy.

In light of its advantage in noise reduction, we use frequencies of \textbf{consecutive 3-gram} entity mentions as the context familiarity proxy.


\begin{table}[t]
    \centering
    \begin{tabular}{c|c}
        \toprule
        \multicolumn{1}{c}{Config} & \multicolumn{1}{c}{AUROC} \\
        \midrule
        \multicolumn{2}{l}{Trivia QA \cite{joshi_triviaqa_2017}} \\
        \midrule
        sentence 2-gram & 60.6 \\
        consecutive 2-gram & 61.8 \\
        \textbf{consecutive 3-gram} & 62.4 \\
        \midrule
        \multicolumn{2}{l}{Natural Questions \cite{joshi_triviaqa_2017}} \\
        \midrule
        sentence 2-gram & 64.2 \\
        consecutive 2-gram & 64.4  \\
        \textbf{consecutive 3-gram} & 64.7  \\
        \bottomrule
    \end{tabular}
    \caption{Correlation between different $Lbl_{corpus}$ configs and $Lbl_{acc}$, measured using area under ROC.}
    \label{tab:lblcorpus}
\end{table}

\section{Hyper-parameters in Classifier Training}
\label{appendix:clsf}

For each MLP classifier, we set the input dimensionality of each layer to be log-linearly decreasing from the LLM hidden size to 64 (at the last MLP layer).

For classifier inputs, we use two popular open-domain QA datasets, TriviaQA \cite{joshi_triviaqa_2017} and Natural Questions (NQ) \cite{kwiatkowski_natural_2019}, where we concatenate the two datasets at each epoch. Since the test sets for both QA datasets have hidden labels, we sub-split the train sets into train2 and dev2 subsets to conduct training, and repurpose the original dev set as a test-set substitute to monitor performance in the training domains.

For NQ dataset, we remove those questions where the gold answers are null. This is because, in the annotation process of the NQ dataset, null gold answers mean the annotators could not find the gold answers from the reference document, where it is unclear whether those answers exist w.r.t. world knowledge, or what those answers would be.

For each QA entry, we take the last layer hidden states at the last question/answer tokens that are \textit{not special tokens (such as </s>).}. We have also experimented using earlier layers and excluding the punctuation tokens, but no improvements have been observed.

We train the faithfulness classifiers with AdamW optimizer \cite{loshchilov2018decoupled}, using learning rate 1e-3. 
We train the classifiers for 300K entries, saving every 25K entries. Using a batch size of 128, this translates to a total of 2344 steps and 196 steps between checkpoint saves. 

Notably, $Lbl_{acc}$ labels are binary, thus we use cross-entropy loss for training; $Lbl_{corpus}$ labels are scalar values between 0 and 1, therefore we use mean-squared error loss for training.
In either case, we track dev2 set losses as the checkpoint selection criterion; when dev2 loss stops decreasing for 2 consecutive saves, we impose early stopping.

\section{Flan-UL2 Results}
\label{appendix:flanul2}

We present hallucination mitigation results with Flan-UL2 in Table \ref{tab:decoding-flanul2}.
We can observe that, as an encoder-decoder model, Flan-UL2 generates responses that are shorter than that from Falcon-7b-instruct. This is related to its fine-tuning on FLAN, a dataset not focused on text generation but classification/reasoning tasks with short responses.

Nevertheless, the results are consistent with Table \ref{tab:decoding-falcon7b}, with H\textsc{allu}C\textsc{ana} outperforming baselines.

Note that corresponding to the generally shorter responses produced by this encoder-decoder LLM, we also relax the length threshold for abstention from 16 words to 8 words.\footnote{By words, we mean space-separated natural language words, a larger unit than LLM tokens.}

\begin{table*}[t]
    \centering
    \begin{tabular}{c|g|ccc|g}
    \toprule
    \textbf{Method} & \boldmath{$GQI_{0.2} \uparrow$} & \textbf{Avg. FS (\%) $\uparrow$} & \textbf{Avg. \#Facts $\uparrow$} & \textbf{\% respond $\uparrow$} & \boldmath{$TpF \downarrow$} \\
    \midrule
    greedy & 0.22 & 21.6 & 7.0 & 99.8\% & 10.2 \\ 
    beam-6 & 0.07 & 24.3 & 6.7 & 98.6\% & 55.6 \\ 
    \textit{MAX($-logp$)} & 0.29 & 22.4 & 6.9 & 93.8\% & 11.0 \\ 
    \textit{SelfCk} \shortcite{manakul_selfcheckgpt_2023} & 0.51 & 27.2 & 7.5 & 63.5\% & 747.9 \\ 
    \midrule
    \multicolumn{6}{l}{\textbf{H\textsc{allu}C\textsc{ana}}} \\
    \cmidrule{2-6}
    $CL_0$ & 0.61 & 29.0 & 6.8 & 57.3\% & 9.9 \\ 
    $CL_x$ & 0.25 & 24.4 & 7.6 & 82.6\% & 58.6 \\ 
    \boldmath{$CL_0+CL_x$} & \textbf{0.59} & \textbf{32.1} & \textbf{7.2} & \textbf{46.1\%} & \textbf{54.0}  \\ 
    
    \bottomrule
    \end{tabular}
    \caption{Hallucination mitigation performance for Flan-UL2 generated biographies for \textbf{factscore test} dataset. For strict computation constraints, $CL_0$ is recommended; generally, ($CL_0 + CL_x$) is recommended.}
    \label{tab:decoding-flanul2}
\end{table*}

\section{Metrics Details}
\label{appendix:metrics}

\subsection{FActScore Implementation Details}
\label{appendix:metrics:factscore}

We follow \citet{min_factscore_2023} to calculate FActScore for generated biographies, w.r.t. Wikipedia as the gold knowledge source. Compared to their original implementation, we made two adjustments. First, we use a more advanced open-source LLM, Mistal \cite{jiang_mistral_2023}, as the backbone LLM, instead of the Inst-LLaMA \cite{touvron_llama_2023,wang_super-naturalinstructions_2022} used in the original implementation. We make this change because the Mistral LLM is independent of the backbone LLMs used in our experiments, exhibits SOTA performance, and is an open-source model with a permissive apache-2.0 licence. Second, instead of penalizing short generations using the $\gamma$ term as in their GitHub release\footnote{\url{https://github.com/shmsw25/FActScore}}, we take into account the generation informativeness in a more fine-grained way, where the $GQI$ metric is proportional to the generation lengths.

In order to verify the validity of the Mistral-based FActScore metric, we compare it with the two variants of FActScores released by \cite{min_factscore_2023} using ChatGPT and Inst-LLaMA. Specifically, we use the LLM generations from the set of LLMs used in their evaluation (Table 3 of their paper), and present the comparison in Table \ref{tab:factscore}. As shown, the ranking among the set of LLMs is consistent between the ``retrieval+Mistral+npm'' setup that we use, and their two original setups, proving its validity.


\begin{table*}[]
    \centering
    \begin{tabular}{c|gcc}
    \toprule
        LLM & retrieval+Mistral+npm & retrieval+ChatGPT & retrieval+llama+npm \\
        \midrule
        InstructGPT & 39.6 & 47.6 & 41.1 \\
        ChatGPT & 56.1 & 65.1 & 58.7 \\
        PPLAI$^\dagger$ & 58.9 & 72.3 & 61.6 \\
        
    \bottomrule
    \end{tabular}
    \caption{FActScores (\%) for the set of 3 LLMs, measured using different configurations; ``retrieval+Mistral+npm'' is the configuration used in our main experiments. Using different configurations, the rankings among the set of LLMs are consistent. PPLAI$^\dagger$ is retieval-augmented.}
    \label{tab:factscore}
\end{table*}

\subsection{Area under FActScore-Rejection Curve}
\label{appendix:metrics:aufrc}

For comparing the discriminative power of different classification approaches on the vanilla greedy LLM generations on the factscore dataset, we use each classifier to rank the generated passages by faithfulness, and compare each ranking against the random ranking and the perfect ranking.

Using the predictions from each classifier, we iteratively reject the passages with the lowest predicted-faithfulness, and calculate an average FActScore supposing that the classifier is thresholded to reject only this and more lowly-predicted passages. We put all the thresholds on a FActScore-RejectionRate plain, where the thresholds are connected as the FActScore-Rejection Curve.

\begin{figure}[ht]
    \centering
    \includegraphics[width=1.0\linewidth]{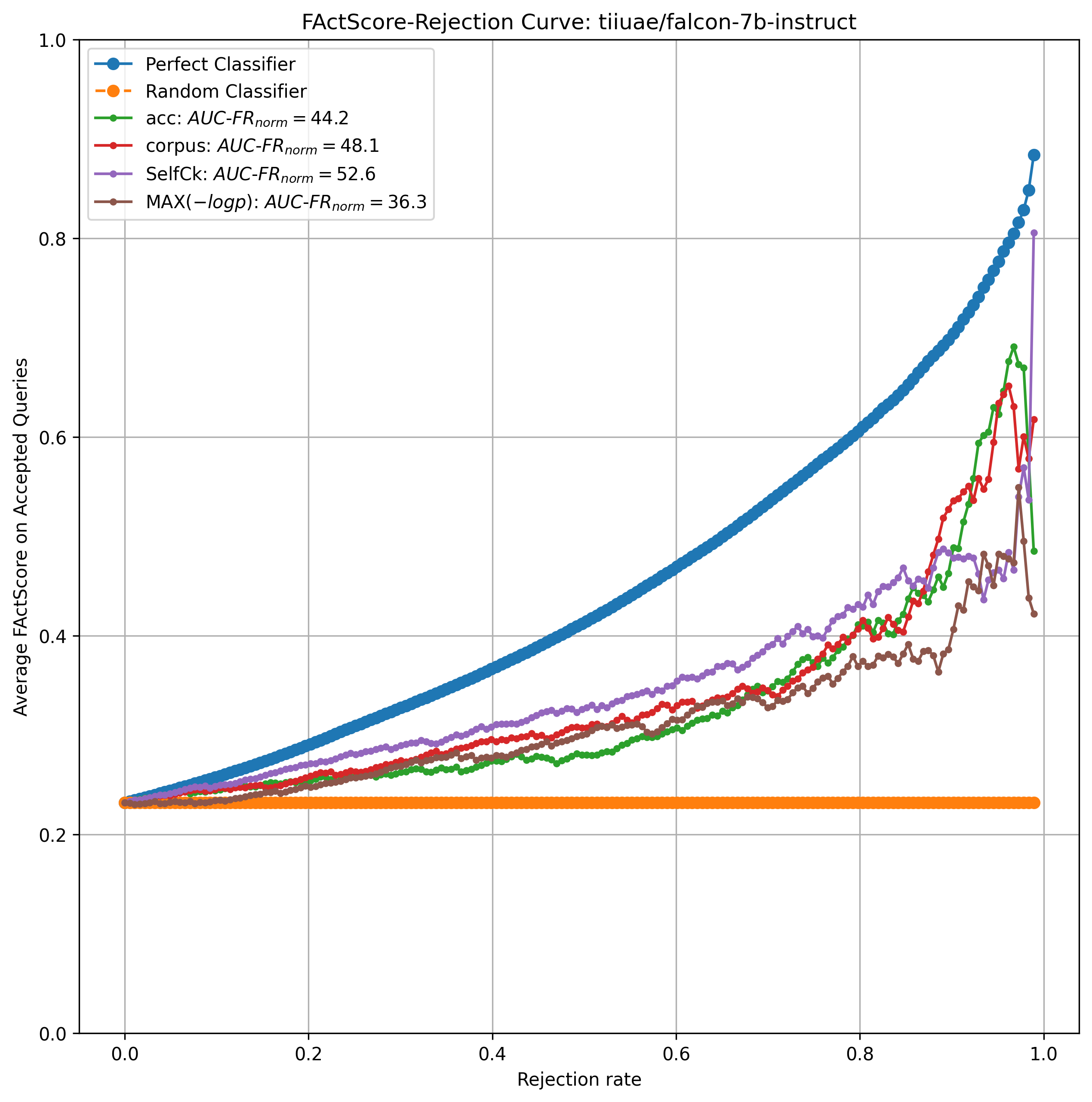}
    \caption{The FActScore-Rejection Curves of various classification approaches, when applied on greedy-decoded generations of Falcon-7b-instruct, over factscore dev set.}
    \label{fig:frc-demo}
\end{figure}

In Figure \ref{fig:frc-demo}, we show an example set of FActScore rejection curves, reported on falcon-7b-instruct greedy generations. The bold \textbf{\textcolor{orange}{orange}} curve represents the random ranking, where precision expectedly does not change no matter how many samples are rejected; the bold \textbf{\textcolor{blue}{blue}} curve represents the perfect ranking, where the lowest FActScore passage is always rejected at each turn.

Actual classifiers have FActScore-Rejection curves between the random and the perfect rankings, so we measure the performance of a classifier by how much above-random discriminative power it has, relative to perfection. This is \boldmath{$\textit{\textbf{AUC-FR}}_{norm}$}, the ratio of above-random area under the FR-curve of the classifier w.r.t. the perfect upper bound.\footnote{On a different classification task of natural language inference, \citet{mckenna_sources_2023} has used an $AUC_{norm}$ metric on the Precision-Recall curve, which is in a similar spirit to the metric we use.}

When selecting a threshold for each classifier to use in the lookahead (\S\ref{sec:experiments:lookahead}), we choose the representative thresholds maximizing F$_\beta$-scores on dev sets. For \textit{classifier}$_0$ classifiers, we set $\beta=1.0$, the default value; for \textit{classifier}$_x$ classifiers, since classification is repeatedly applied throughout a generation run, and vetoing all options at any critical time step leads to generation being halted, we use $\beta=1.2$ to avoid over-rejection.

\subsection{Generation Quality Index ($GQI$)}
\label{appendix:metrics:gqi}

We provide $GQI$ as an intuitive holistic metric of LLM generation quality, using one single index. This metric is provided for the reader's convenience, and reflects how we value a generated response, taking into account factuality, informativeness and responsiveness:

\begin{equation*}
    gqi_{\gamma}(d) = (\textit{FS}_d-\gamma)\times \textit{\#Facts}_d
\end{equation*}
\begin{equation*}
    GQI_{\gamma}=\frac{\sum_{d\in D}gqi_{\gamma}(d)}{|D|\times \sqrt{\textit{response rate}}}
\end{equation*}

The design of $GQI$ follows the tenets below:

First, responses with $FS_d<\gamma$ are negative assets, as they do not provide useful information and only add to the confusion. Relative to the baseline performance of the LLMs we test, we empirically set $\gamma=0.2$.

Second, for the same total number of atomic facts, more informative responses for fewer entries are preferred, over more but less informative responses; this is because, while nationalities etc., which come at the beginning of a biography, are easier to infer from the names etc., later generations are more knowledge-intensive. 
In the denominator term, taking $(\textit{response rate})^0 \equiv 1$ means responsiveness is taken equally as informativeness; on the other hand, taking $(\textit{response rate})^1 \equiv \textit{response rate}$ means responsiveness is completely ignored. We take a balance between the two extremes and take the square root $\sqrt{\cdot}$.


We acknowledge that the $GQI$ metric reflects our beliefs in the value of LLM long-form generations, and may not universally apply. We encourage readers to also inspect the individual measures of each aspect to gain a complete view and make judgements w.r.t. their own applications.

\begin{table*}[h]
    \centering
    \begin{tabular}{c|p{2.71in}|p{2.71in}}
    \toprule
        \textbf{Index} & \multicolumn{1}{P{2.71in}}{\textbf{Greedy}} & \multicolumn{1}{|P{2.71in}}{\textbf{H\textsc{allu}C\textsc{ana}}} \\
        \midrule
        3 & Iggy Azazela is a multi-platinum, Australian rapper, singer, songwriter, and actress. She rose to fame in 2010 with the release of her debut single, 'Iggy Azalea,' which reached number one on the Australian Singles Chart. Since then, she has released several successful albums, including 'Reign,' 'The New Classic,' and 'In My Defense.' Azalea has also been featured on songs by other artists, including Eminem and Ariana Grande. She has been nominated for multiple Grammy Awards and has won multiple awards for her music. In addition to her music career, Azalea has also appeared in movies and TV shows, including 'Furious 7' and 'The Fast and Furious' franchise. & Iggy Azazela is a rapper, singer, and songwriter from Australia. She gained popularity in 2010 with the release of her mixtapes Iggy Azalea and Reclassified. Azalea rose to fame with the release of her debut studio album, The New Classic, in 2014. She has since been nominated for numerous awards, including multiple ARIA Awards and a BET Award. \\
        \midrule
        6 & Radhika is an Indian actress who has appeared in numerous Hindi, Telugu, and Tamil films. She has also acted in a few English-language films. She is known for her roles in the web series 'The Raikar Case' and the film 'The Perfect Girl 2.' Radhika has won several awards for her performances, including the Filmfare Award for Best Actress in a Supporting Role. & Radhika is an Indian actress who has predominantly worked in Telugu and Tamil films. She has also appeared in a few Hindi films. She has been a part of critically acclaimed films like Raja Reddy, RRR, and Jersey. She has won several awards for her performances, including the Filmfare Award for Best Actress in a Supporting Role \textcolor{purple}{for.} She is known for her versatile roles and is considered one of the most sought-after actresses in the South Indian film industry. \\
        \midrule
        7 & David Galloway is a writer and journalist based in London. He has written for a variety of publications, including The Guardian, The Independent, and The New York Times. He is the author of the book 'The New York Times Best of 2008' and has written for the BBC, The Economist, and The Financial Times. He is currently working on a book about the history of the New York Stock Exchange. & David Galloway is a freelance writer and editor based in the UK. He specializes in writing about travel, food, and culture. He has been writing professionally for over 10 years \textcolor{purple}{and has.} \\
    \bottomrule
    \end{tabular}
    \caption{Comparing biographies generated by Falcon-7b-instruct using greedy search and H\textsc{allu}C\textsc{ana}. Examples are the first 3 entries in factscore test set where both responses are not abstained, indices are from the original test set.}
    \label{tab:case_study}
\end{table*}

\section{Qualitative Analysis on Generation Fluency}
\label{appendix:case_study}

In Table \ref{tab:case_study}, we compare the responses generated using vanilla greedy decoding, against those generated using H\textsc{allu}C\textsc{ana}. We present the first 3 entries in factscore test set where both responses are not abstained. Since the performance of H\textsc{allu}C\textsc{ana} on hallucination reduction has been quantitatively verified in the above, in this section we are primarily interested in understanding how well it has preserved the fluency of the LLM generations.

We can observe that responses generated using HalluCana remain fluent and grammatical, as much so as the greedy decoding. The only caveats come at the 2 \textcolor{purple}{purple} coloured text spans: these are when the veto mechanism rejects all the likely branches, and we force the next token to be period (.) to avoid hallucination.

Since these spans do not significantly impact our quantitative evaluations, we do not specifically take a stance on how to post-process the output to remove these spans. We do offer a few possible options, including: 1) deleting words in these spans with forced periods, until the sentence is syntactically complete, e.g. with a lightweight parser; 2) marking ``[UNK]'' for the slots; 3) querying knowledge sources or larger LLMs at these points to fill in these blanks. 

Additionally, we note that a back-tracking approach in LLM decoding has been proposed \cite{liu_correction_2023}. It is also possible to avoid the force-stopped sentences by integrating the back-tracking approach in our algorithm. We leave these explorations to future work.

\section{Computational Infrastructure and Footprint}
\label{appendix:compute}

For the experiments in this paper, we use a g5-12xlarge cloud server from AWS\footnote{\url{https://aws.amazon.com/ec2/instance-types/g5/}}. Training faithfulness classifiers consumes ~15 minutes using only 1 GPU. Generating long-form responses consumes ~30 minutes on factscore dev-set and ~70 minutes on factscore test set; 2 A10G GPUs are required for generating responses using Falcon-7B, on the other hand, all 4 A10G GPUs on the server are required for generating responses using Flan-UL2. Evaluating generated responses with Mistral-based FActScore metric consumes ~2 hours for factscore dev-set, and ~5 hours for factscore test-set, requiring 2 A10G GPUs.

The most computationally demanding part of the pipeline is obtaining linked entities for LLM pre-train corpora using ReFinED entity linker, which consumes ~3000 GPU hours for the C4 corpus, and ~6000 GPU hours for Falcon-ReFinedWeb. We refer readers to \citet{kandpal_large_2023} for a rough substitute, where C4 corpus is parsed using an older entity linker \cite{mendes_dbpedia_2011}.

\end{document}